\documentclass[twoside,twocolumn,12pt]{article}
\usepackage{extsizes}
\usepackage[super,sort&compress,comma]{natbib} 
\usepackage[version=3]{mhchem}
\usepackage[left=1.5cm, right=1.5cm, top=1.785cm, bottom=2.0cm]{geometry}
\usepackage{balance}
\usepackage{times,mathptmx}
\usepackage{sectsty}
\usepackage{graphicx} 
\usepackage{lastpage}
\usepackage[format=plain,justification=justified,singlelinecheck=false,font={stretch=1.125,small,sf},labelfont=bf,labelsep=space]{caption}
\usepackage{float}
\usepackage{fancyhdr}
\usepackage{fnpos}
\usepackage[english]{babel}
\addto{\captionsenglish}{%
  
}
\usepackage{array}
\usepackage{droidsans}
\usepackage{charter}
\usepackage[T1]{fontenc}
\usepackage[usenames,dvipsnames]{xcolor}
\usepackage{setspace}
\usepackage[compact]{titlesec}
\usepackage{hyperref}
\usepackage{amsfonts}
\usepackage{amssymb}

\usepackage{epstopdf}

\definecolor{cream}{RGB}{222,217,201}

\begin{document}

\makeFNbottom
\makeatletter
\renewcommand\LARGE{\@setfontsize\LARGE{15pt}{17}}
\renewcommand\Large{\@setfontsize\Large{12pt}{14}}
\renewcommand\large{\@setfontsize\large{10pt}{12}}
\renewcommand\footnotesize{\@setfontsize\footnotesize{7pt}{10}}
\makeatother

\renewcommand{\thefootnote}{\fnsymbol{footnote}}
\renewcommand\footnoterule{\vspace*{1pt}%
\color{cream}\hrule width 3.5in height 0.4pt \color{black}\vspace*{5pt}} 
\setcounter{secnumdepth}{5}

\makeatletter 
\renewcommand\@biblabel[1]{#1}            
\renewcommand\@makefntext[1]%
{\noindent\makebox[0pt][r]{\@thefnmark\,}#1}
\makeatother 
\renewcommand{\figurename}{\small{Figure}~}
\sectionfont{\sffamily\Large}
\subsectionfont{\normalsize}
\subsubsectionfont{\bf}
\setstretch{1.125} 
\setlength{\skip\footins}{0.8cm}
\setlength{\footnotesep}{0.25cm}
\setlength{\jot}{10pt}
\titlespacing*{\section}{0pt}{4pt}{4pt}
\titlespacing*{\subsection}{0pt}{15pt}{1pt}

\renewcommand{\footrulewidth}{0pt}
\setlength{\arrayrulewidth}{1pt}
\setlength{\columnsep}{6.5mm}
\setlength\bibsep{1pt}

\makeatletter 
\newlength{\figrulesep} 
\setlength{\figrulesep}{0.5\textfloatsep} 

\newcommand{\topfigrule}{\vspace*{-1pt}%
\noindent{\color{cream}\rule[-\figrulesep]{\columnwidth}{1.5pt}} }

\newcommand{\botfigrule}{\vspace*{-2pt}%
\noindent{\color{cream}\rule[\figrulesep]{\columnwidth}{1.5pt}} }

\newcommand{\dblfigrule}{\vspace*{-1pt}%
\noindent{\color{cream}\rule[-\figrulesep]{\textwidth}{1.5pt}} }

\makeatother

\twocolumn[
  \begin{@twocolumnfalse}
    \vspace{0.6cm}
    \sffamily
    \centering\LARGE{\textbf{SmoothQuant+: Accurate and Efficient 4-bit Post-Training Weight Quantization for LLM}} \\
    \vspace{0.6cm}

    \large{Jiayi Pan, Chengcan Wang, Kaifu Zheng, Yangguang Li, Zhenyu Wang, Bin Feng} \\
    \large{ZTE Corporation}

 \end{@twocolumnfalse} \vspace{0.6cm}

]
  

\renewcommand*\rmdefault{bch}\normalfont\upshape
\rmfamily
\section*{}
\vspace{-1cm}

\section*{\centering{Abstract}}
Large language models (LLMs) have shown remarkable capabilities in various tasks. However their huge model size and the consequent demand for computational and memory resources also pose challenges to model deployment. Currently, 4-bit post-training quantization (PTQ) has achieved some success in LLMs, reducing the memory footprint by approximately 75\% compared to FP16 models, albeit with some accuracy loss. In this paper, we propose SmoothQuant+, an accurate and efficient 4-bit weight-only PTQ that requires no additional training, which enables lossless in accuracy for LLMs for the first time. Based on the fact that the loss of weight quantization is amplified by the activation outliers, SmoothQuant+ smoothes the activation outliers by channel before quantization, while adjusting the corresponding weights for mathematical equivalence, and then performs group-wise 4-bit weight quantization for linear layers. We have integrated SmoothQuant+ into the vLLM framework, an advanced high-throughput inference engine specially developed for LLMs, and equipped it with an efficient W4A16 CUDA kernels, so that vLLM can seamlessly support SmoothQuant+ 4-bit weight quantization. Our results show that, with SmoothQuant+, the Code Llama-34B model can be quantized and deployed on a A100 40GB GPU, achieving lossless accuracy and a throughput increase of 1.9 to 4.0 times compared to the FP16 model deployed on two A100 40GB GPUs. Moreover, the latency per token is only 68\% of the FP16 model deployed on two A100 40GB GPUs. This is the state-of-the-art 4-bit weight quantization for LLMs as we know. Code$\dag$ is available at https://github.com/adlik/smoothquant+, https://github.com/adlik/vllm.

\footnotetext{\dag~https://github.com/adlik/smoothquant+ to be released in the near future}

\section{Introduction}
Generative pre-trained language models, evolving from the transformer architecture\cite{vaswani2017attention}, are getting larger and larger, from GPT-2\cite{radford2019language} with 1.5B parameters to GPT-NeoX\cite{black2204gpt} with 20B parameters, and then to GPT-3\cite{brown2020language} with 175B parameters. The number of parameters of PaLM\cite{chowdhery2022palm} has further reached 540B. The increasing number of parameters grants these LLMs enhanced capabilities in handling some complex tasks. The advent of models like ChatGPT has illuminated the new direction in AI development. However, as the models grow larger, these LLMs often encounter various challenges when deployed such as high latency, significant computing resource consumption, and large memory footprints become more pronounced. For example, GPT-3 developed by OpenAI in 2020 contains 175 billion parameters. It requires 700GB of memory for operation in FP32, and still a substantial 350GB in FP16. This necessitates the use of at least five high-end GPUs, like NVIDIA's A100 80GB, for inference, which greatly limits the scope of use of the models. 

Quantization is an effective way to reduce memory footprint and computing resource overhead when deploying LLMs. At present, 8-bit quantization of LLMs is relatively mature\cite{dettmers2022llm,xiao2023smoothquant}, the accuracy can basically not be degraded while the memory footprint is halved. 4-bit weight quantization can further lower the service threshold of LLMs, only 1/4 memory footprint is needed. However, it still faces performance challenges. The quantization process is time-consuming and the performance loss after quantization is large, whether it's through quantization-aware training (QAT) methods like LLM-QAT\cite{liu2023llm}, or post-training quantization (PTQ) approaches like GPTQ\cite{frantar2022gptq} and AWQ\cite{lin2023awq}. GPTQ, for instance, employs second-order information for error compensation but risks overfitting to the calibration set. AWQ considers that weights with different amplitudes are of different importance to LLM performance. Protecting important weights by adjusting the amplitude of input activation can reduce quantization losses. However, AWQ searches scaling factors layer by layer, the problem of error accumulation is not taken into account during the searching process, which makes the search speed slow and the model accuracy degraded. For example, the HumanEval accuracy is degraded by 4.3\% for the Code Llama-13B quantized by AWQ.

Through investigation, we believe that the weight distribution of the model is uniform and flat, easy to quantize, but the activation distribution fluctuates greatly. Systematic outliers occur beyond 6.7B scale. Outliers appear in patches on the same channels of different tokens. These outliers lead to the difficulty of activation quantification\cite{dettmers2022llm,xiao2023smoothquant}. Even if only 4-bit weight quantization is performed, the model accuracy decreases significantly\cite{frantar2022gptq,lin2023awq,lee2023owq}. We believe that this is caused by the loss of weight quantization being amplified by the activation outliers. In this paper, we propose SmoothQuant+, an accurate and efficient 4-bit weight quantization. SmoothQuant+ minimizes the quantization loss of the model by smoothing activation outliers and adjusting corresponding weights to reduce the impact of activating outliers on the overall quantization loss. We conduct a series of experiments on the Code Llama family\cite{roziere2023code}. SmoothQuant+ achieves lossless 4-bit weight quantization on Code Llama-7B, 13B, and 34B models. It is worth mentioning that Code Llama-34B in FP16 originally needs two NVIDIA A100 40GB GPUs for deployment. By using SmoothQuant+, only a single GPU is needed. The throughput is 1.9-4.0 times that of the model in FP16 deployed on two A100 40GB GPUs, and the latency per token is only 68\% of the model in FP16.
\vspace{-2mm}
\section{SmoothQuant+}
\subsection{Preliminaries}
Quantization maps high-precision floating point numbers to low-precision integers. $\mathbf{W}^{\mathrm{FP} 16} $ is the floating point weight. $\overline{\mathbf{W}}^{\mathrm{INT} 4}$ is obtained by 4-bit uniform quantization. $\hat{\mathbf{W}}^{\mathrm{FP} 16}$ is the quantized weight. The quantization process is as follows:
\begin{align}\label{eq1}
& \overline{\mathbf{W}}^{\mathrm{INT} 4}=\text { clamp }\left(\text { round }\left(\frac{\mathbf{W}^{\mathrm{FP} 16}}{\Delta}\right)+\mathbf{Z}, 0,2^N-1\right) \notag,\\
& \hat{\mathbf{W}}^{\mathrm{FP} 16}=\left(\overline{\mathbf{W}}^{\mathrm{INT} 4}-\mathbf{Z}\right) \times \Delta, \\
& \Delta=\frac{\mathbf{W}_{\max }^{\mathrm{FP} 16}-\mathbf{W}_{\min }^{\mathrm{FP} 16}}{2 ^N-1} \notag.
\end{align}
Here, $\mathbf{Z}$ is zero point, $\Delta$ is the quantization step size,  $clamp \left(z,r_1,r_2\right)$ returns z with values below $r_1$ set to $r_1$ and values above $r_2$ set to $r_2$. round(z) rounds z to the nearest integer. The quantization error is mainly introduced by the round(z) function. The straight through estimator(STE)\cite{bengio2013estimating} is usually used to approximate the gradient through the round function as a pass through operation.
For a linear layer in Transformers\cite{vaswani2017attention}:
\begin{equation}\label{eq2}
\begin{aligned}
& \mathbf{Y}=\mathbf{X W}^{\mathrm{FP} 16}, \\
& X \in \mathbb{R}^{T \times C_i}, \mathbf{W} \in \mathbb{R}^{C_i \times C_o}, \mathbf{Y} \in \mathbb{R}^{T \times C_o} .
\end{aligned}
\end{equation}

By quantizing the weight and substitute formula \eqref{eq1} into formula \eqref{eq2}, we get:
\begin{align}\label{eq3}
\mathbf{Y}=\mathbf{X} \hat{\mathbf{W}}^{\mathrm{FP16}}.
\end{align}
The quantization loss is as follows:
\begin{align}\label{eq4}
\mathrm{E}=\left\|\mathbf{X W}^{\mathrm{FP} 16}-\mathbf{X} \hat{\mathbf{W}}^{\mathrm{FP} 16}\right\|_2^2
\end{align}

\subsection{Smoothing LLMs}
When the number of model parameters of LLMs exceeds 6.7B, systematic outliers appear in activation, which leads to an increase in quantization error and a decrease in accuracy\cite{dettmers2022llm}. To address this problem, LLM.int8() adopts a mixed-precision decomposition (outliers and their corresponding weights are calculated using FP16, and the others are calculated in INT8). These activation outliers appear on almost all tokens but are limited to fixed channels in the hidden state dimension. Given a token, the variances between different channels are large, while the variance is small for different tokens within the same channel. Considering that these outliers in activations are often 100 times larger than most of the activation values, the task of activation quantization becomes particularly challenging. Therefore, SmoothQuant\cite{xiao2023smoothquant} proposes a mathematically equivalent per-channel scaling transformation, which aims to smooth the activations of different channels and their corresponding weights, making the model more friendly to quantization.

\begin{figure}
 \centering
 \includegraphics[height=7cm]{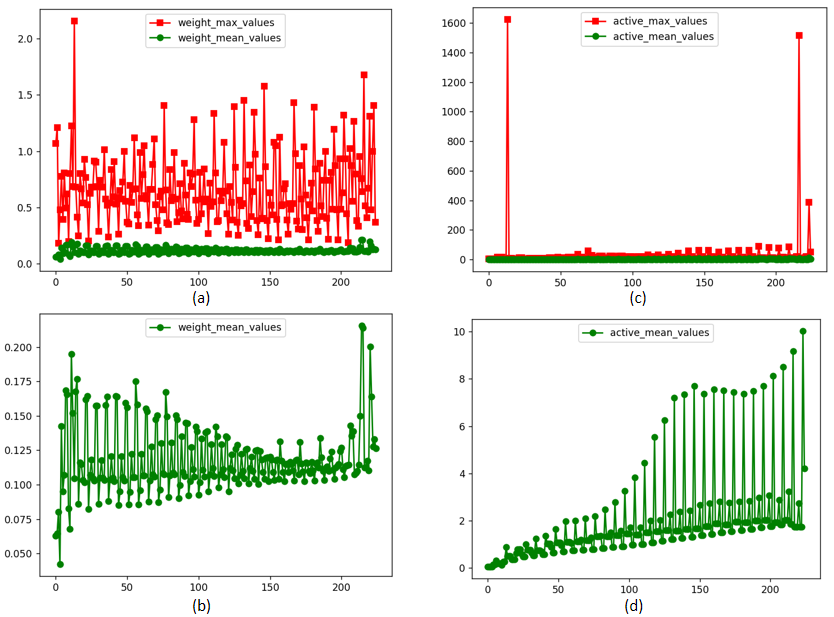}
 \caption {All weights and activation distributions of linear layers in the Code Llama-7B model (taking a subset of the Pile validation set\cite{gao2020pile} as input ). The horizontal axis is the index of the linear layer, and the vertical axis is the absolute value of the amplitude. It is seen from the figure that the activation distribution is uneven and the fluctuation of the activation is significantly larger than the weight. The mean value of the weight is below 0.3, and the maximum value is below 2.5. The mean value of the activation can reach a maximum of 10, and the maximum value can reach 1600.}
 \label{fgr:weights_act_distributions}
\end{figure}

\begin{figure*}[!ht]
 \centering
 \includegraphics[height=8cm,keepaspectratio]{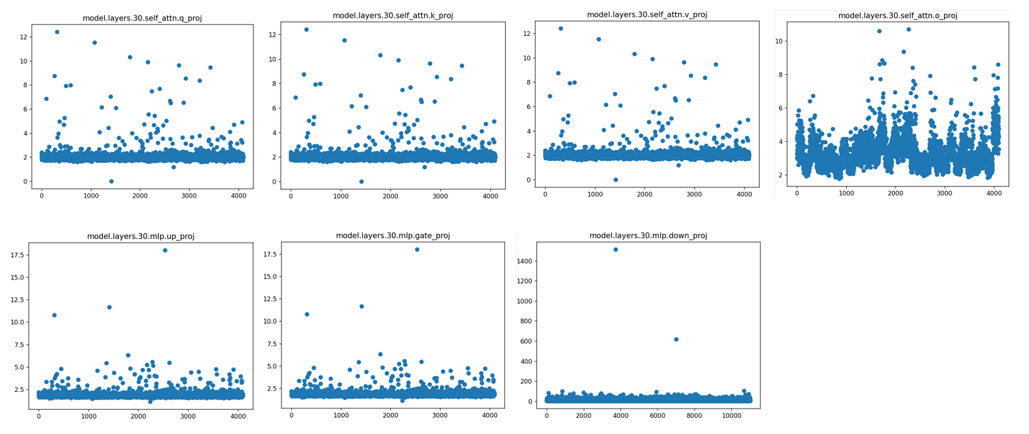}
 \caption {Activation analysis of 7 linear layers of the LlamaDecoderLayer (model. layers. 30) in Code Llama-7B. The horizontal axis is the channel dimension in activation, and the vertical axis is the absolute value of the amplitude. It use a subset of the Pile validation set\cite{gao2020pile} as model input. From the figure, it can be seen that the activation outliers always appear in a small number of fixed channels. The amplitudes of these activation outliers can reach 100 times that of other activation amplitudes.}
 \label{fgr:activation_analysis}
\end{figure*}

\begin{figure}[!ht]
 \centering
 \includegraphics[height=6cm,width=9cm,keepaspectratio]{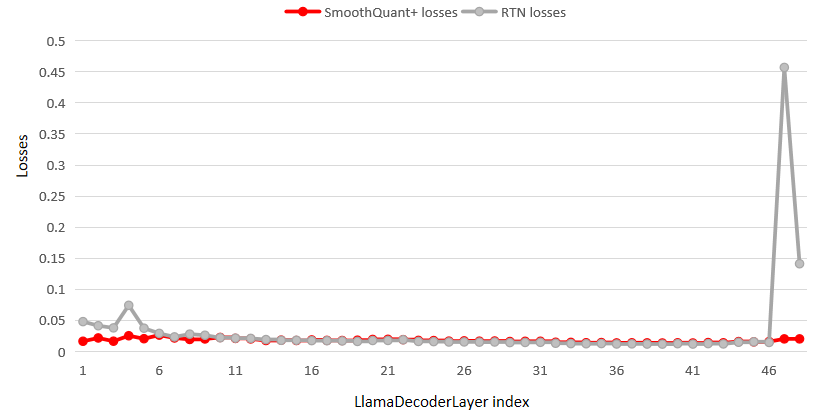}
 \caption {Comparison of quantization losses for each LlamaDecoderLayer in Code Llama-7B.}
 \label{fgr:compare_quant_losses}
\end{figure}

Figure \ref{fgr:weights_act_distributions}(a)(b) illustrates that the magnitude of the weights is evenly distributed, suggesting that it should be easy to quantize. Nevertheless, methods such as  GPTQ\cite{frantar2022gptq}, AWQ\cite{lin2023awq}, or OWQ\cite{lee2023owq} performing 4-bit weight quantization still result in some loss of accuracy. Currently, we observe that outliers in activation always appear in a small number of fixed channels. If a channel has an outlier, it tends to persist in all tokens. The amplitudes of these outliers are often 100 times greater than other activation amplitudes\cite{xiao2023smoothquant}. This can be confirmed from Figure \ref{fgr:weights_act_distributions} and Figure \ref{fgr:activation_analysis}. According to the formula \eqref{eq4}, the quantization loss is not only related to the weight, but also to the activation, with the two beings in a multiplicative relationship. This suggests that the loss of weight quantization is amplified by the activation outliers. By smoothing the activation outliers and adjusting the weights accordingly, the quantization loss can be greatly reduced. Figure \ref{fgr:compare_quant_losses} is a comparison of the quantization losses of each LlamaDecoderLayer in Code Llama-7B. It shows that smoothing the model before quantization can flatten the loss peaks and significantly reduce the quantization loss.

\begin{figure}[!ht]
 \centering
 \includegraphics[height=6cm,width=10cm,keepaspectratio]{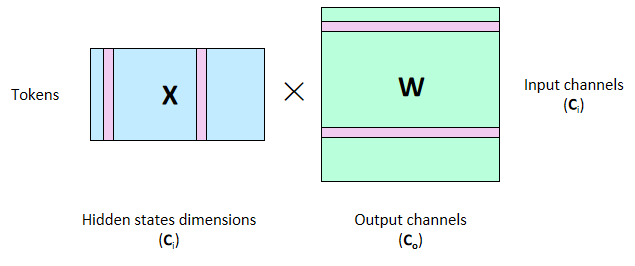}
 \caption {Smoothing process of SmoothQuant+. Pink indicates smoothing operations by channels. Activations are smoothed along the dimension of the hidden states, and weights are smoothed over the dimension of the input channels.}
 \label{fgr:smoothing_process}
\end{figure}

Based on the above considerations, we consider smoothing the activations like SmoothQuant\cite{xiao2023smoothquant}, dividing by the smoothing factor $\text{s}\in \mathbb{R}^{C_i}$ by channel. To maintain mathematical equivalence of linear layers, the weights are adjusted accordingly in the opposite way:
\begin{align}\label{eq5}
\mathbf{Y}=\left(\mathbf{X}\text{diag}(\mathbf{s})^{-1}\right)\cdotp(\text{diag}(\mathbf{s})\mathbf{W})=\hat{\mathbf{X}}\hat{\mathbf{W}}.
\end{align}
The process of smoothing activation $\mathbf{X}$ and weight $\mathbf{W}$ by channel is shown in the Figure \ref{fgr:smoothing_process}.

\begin{figure}
 \centering
 \includegraphics[height=8cm]{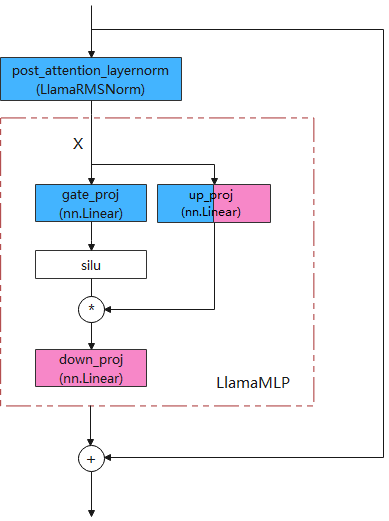}
 \caption {Smoothing LlamaMLP in Code Llama. The same color block indicates using the same smoothing factor.}
 \label{fgr:smoothing_mlp}
\end{figure}

Considering that the activation $\mathbf{X}$ is usually generated by linear operations in the previous layer (such as linear layers, layer norms, etc.), it can easily fuse the smoothing factor into the parameters of the previous layer as well as residual connection. Figure \ref{fgr:smoothing_mlp} takes the LlamaMLP structure as an example to illustrate the fusing process of the residual branch briefly. In Figure \ref{fgr:smoothing_mlp}, the same color block indicates using the same smoothing factor. In the LlamaMLP of Code Llama, there are 3 linear layers, where gate\_proj and up\_proj have the same input $\mathbf{X}$ which is the output of LlamaRMSNorm named post\_attention\_layernorm. Dividing $\mathbf{X}$ by the smoothing factor can be integrated into post\_attention\_layernorm’s parameters. To keep the mathematical equivalence, the weights of gate\_proj and up\_proj are multiplied by the smoothing factor, respectively. For down\_proj, only up\_proj is a linear operation in front of it, so the operation of dividing its input by the smoothing factor is fused into the weights of up\_proj, and the weights of down\_proj are multiplied by the smoothing factor to maintain the mathematical equivalence of the model.

It is necessary to choose an appropriate smoothing factor $\mathbf{s}$ to smooth activation by channel, i.e. $\mathbf{Y}=\mathbf{X}\text{diag}(\mathbf{s})^{-1}$. A straight-forward choice is $\mathbf{s}_j=\max \left(\left|\mathbf{X}_j\right|\right), j=1,2, \ldots, \mathrm{C}_i$, where $j$ corresponds to $j$-th input channel. By division, all activation channels have the same maximum value of 1. The amplitude fluctuation of the weight is small, so it is easy to quantize. It reachs a state that the weight quantization loss can not be differentially amplified by activation.  However, given that the activation range is dynamic and the research primarily focuses on the quantization's impact on code generation, we use the problem descriptions in HumanEval\cite{chen2021evaluating} for calibration. Nonetheless, to maintain mathematical equivalence, the weights of the linear layer need to be adjusted by channel, which increases the fluctuation of the weights, leading to larger quantization error. From the perspective of easy quantization of weights, we hope to choose $\mathbf{s}_j=1/\max\left(\left|\mathbf{W}_j\right|\right)$. However, this in turn makes the activation fluctuations greater. So we choose $\mathbf{s}$, which can smooth the activations without making the weights difficult to quantify, where smoothing strength $\alpha$ controls the strength of smoothing the activations according to formula \eqref{eq6}. At the same time, in order to further reduce the quantization loss, a finer quantization granularity is selected, using group-wise quantization. Group-size is usually set to be 128.

\begin{align}\label{eq6}
\mathbf{s}_j=\max\left(\left|\mathbf{X}_j\right|\right)^\alpha/\max\left(\left|\mathbf{W}_j\right|\right)^{1-\alpha}
\end{align}

During the smoothing process, it is crucial to select an appropriate smoothing strength $\alpha$. We use grid search with an interval of 0.05 between 0 and 1 to search for the smoothing strength that minimizes the quantization loss of the entire model. In this process, 164 problem descriptions in HumanEval\cite{chen2021evaluating} are used as a calibration set.

\footnotetext{\dag~https://github.com/vllm-project/vllm}

\subsection{4-bit Quantization in vLLM}
vLLM$^\dag$ is a high-throughput and memory-efficient inference and serving engine for LLMs. At present, quantification in vLLM supports AWQ and needs to be developed for different models by the end users. As far as we know, only LLaMA models are originally supported by vLLM, and we believe that this implementation of quantization is not friendly to the architecture of vLLM itself. Before quantization, it is necessary to generate a quantified model using AWQ\cite{lin2023awq}$^\ddag$ and then load it into vLLM.

\footnotetext{\ddag~https://github.com/casper-hansen/AutoAWQ}

We implement an efficient 4-bit group-wise quantization specifically for vLLM architecture. The process begins by loading the smoothed FP16 model into vLLM. Initially, the model is loaded into memory, and then the weights of linear layers are group-wise quantized with 4-bit during the process of migrating the model from CPU to GPU. Consequently, the model that eventually resides in the GPU memory is already quantized to 4 bits. During inference, SmoothQuant+ uses the efficient W4A16 CUDA kernel optimized from LMDeploy$^\S$ for the quantized linear layers. For the Code Llama-34B, only a single A100 40GB GPU is needed. Both in terms of throughput and latency, it comprehensively surpasses FP16 model deployed on two GPUs in performance.

\footnotetext{\S~https://github.com/InternLM/lmdeploy}

The benefits of implementing quantization in our way on vLLM are:
\begin{itemize}
    \item User-Friendly Deployment: Users do not need to differentiate between deploying an FP16 model or a quantized model. All models loaded are the original unquantized FP16 models from Huggingface, simplifying the user experience. 
    \item Seamless Integration with vLLM: The method is designed to be compatible with vLLM, requiring no special consideration for model differences. Any model supported by vLLM can seamlessly support group-wise quantization, including SmoothQuant+.
    \item Versatility in Group Sizes: Support group-wise quantization for different group sizes.
    \item Efficient Resource Utilization: We not only reduce the memory footprint of LLMs to 1/4 of FP16’s, but also increase the throughput during inference and reduce inference latency.
\end{itemize}

\begin{figure}[!ht]
 \centering
 \includegraphics[height=10cm,width=8cm,keepaspectratio]{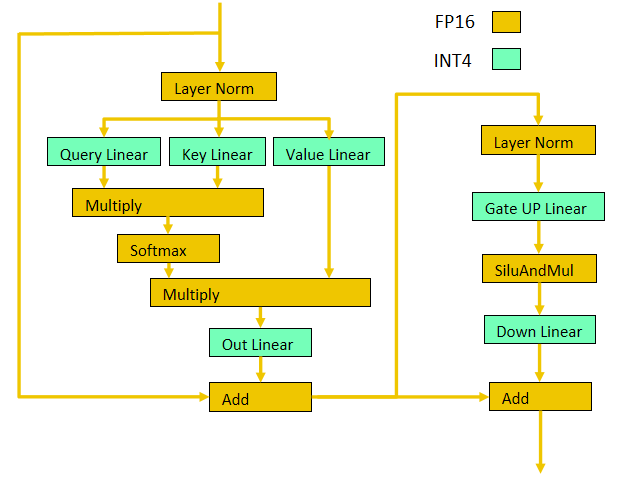}
 \caption {Precision mapping of LlamaDecoderLayer in Code Llama. Yellow indicates that the data exists in FP16, and green indicates that the data exists in INT4. Since the W4A16 core is used during inference, the input and output of all linear layers in the figure are FP16 during inference.}
 \label{fgr:precision_mapping}
\end{figure}

The precision mapping of the LlamaDecoderLayer in Code Llama using SmoothQuant+ is shown in Figure \ref{fgr:precision_mapping}.

\footnotetext{\dag~https://github.com/google-research/babelcode}

\section{Experiments}
\subsection{Settings}
\subsubsection{Models and Datasets} 
We choose Code Llama family\cite{roziere2023code} to evaluate SmoothQuant+, focusing on evaluating the impact of quantization on code generation. For Python, HumanEval\cite{chen2021evaluating} is used as the evaluation benchmark. The BabelCode framework\cite{orlanski2023measuring}$^\dag$ which converts the HumanEval (Python) set into other programming languages, including Python, JAVA, Go, C++, is used for multilingual evaluation.

\subsubsection{Baselines} 
To evaluate SmoothQuant+, our experiments include three baselines: one is the unquantized FP16 model, which is our main comparison scenario, because we hope that the model after 4-bit weight quantization is lossless. The other is round-to-nearest (RTN) quantization, which quantizes the weights 4-bit group-wise with group-size 128. RTN is a benchmark for evaluating whether a quantization method is more effective than the ordinary method. We also compare with the state-of-the-art 4-bit weight quantization method AWQ\cite{lin2023awq}.

\subsubsection{Implementation}
All the tests are completed on the vLLM inference engine, including tests on FP16 models, RTN, AWQ, and SmoothQuant+. RTN and SmoothQuant+ are implemented by ourselves on vLLM, while AWQ is already supported by the vLLM framework.

The SmoothQuant+ algorithm smoothes the model by channel, and then performs 4-bit group-wise quantization on the weights of the model, typically with a group size of 128. The methods are as follows:
\begin{itemize}
    \item \pmb{Smoothing:} When smoothing, it is crucial to select an appropriate smoothing strength $\alpha$. We search for the smoothing strength that minimizes the quantization loss of the entire model at an interval of 0.05 between 0 and 1. In this process, 164 problem descriptions in HumanEval\cite{chen2021evaluating} are used as the calibration set. The search process is very fast which only requires 1/5 of the time taken by AWQ for the Code Llama-34B model. After searching for a suitable one, we smooth the model according to formula \eqref{eq5}.
    \item \pmb{Quantization}: We implement 4-bit group-wise quantization (RTN) for linear layers on vLLM. The smoothed model is directly loaded into vLLM, which automatically completes 4-bit weight quantization. In vLLM, we have implemented an efficient W4A16 CUDA kernel optimized from LMDeploy for quantization of linear layers, which further enhances the acceleration effect. By using SmoothQuant+,  the Code Llama-34B can be quantified and deployed on a single A100 40GB GPU, with lossless accuracy and a throughput increase of 1.9-4.0 times compared to the FP16 model deployed on two A100 40GB GPUs. The latency per token is only 68\% of the FP16 model deployed on two A100 40GB GPUs. This is the state-of-the-art 4-bit weight quantization as we know.
\end{itemize}

\subsection{Accuracy Evaluation}
\subsubsection{Python Code Generation}
We focus on studying the impact of 4-bit weight quantization on the Code Llama family for its model structure is roughly the same as that of the open-source LLaMA models, except for minor changes in the implementation of the attention layer. The subsequent WizardCoder\cite{luo2023wizardcoder} with better code generation capabilities is also fine-tuned based on the Code Llama family. We evaluate Python code generation capabilities of Code Llama family under FP16, RTN, AWQ, and SmoothQuant+ in Table \ref{tbl:humaneval_python}. It can be seen that SmoothQuant+ perfectly achieves lossless 4-bit weight-only quantization under HumanEval-Python. For the Code Llama-34B model, the accuracy is not only 2.44\% higher than AWQ, but also 1.83\% higher than the unquantized FP16 model. AWQ cannot achieve lossless 4-bit weight-only quantization under the vLLM framework. For Code Llama-13B, AWQ has a 4.27\% lower accuracy than the FP16 model, and even for the 34B model, the accuracy is 0.61\% lower.

\begin{table}
\small
  \caption{Code Llama family HumanEval-Python pass@1 on vLLM engine}
  \label{tbl:humaneval_python}
  \setlength{\tabcolsep}{3mm}{
  \begin{tabular*}{\textwidth}{lccc}
    \cline{1-4}
    HumanEval$\uparrow$ & 7B & 13B & 34B \\
    \cline{1-4}
    FP16 & 35.98\% & 35.98\% & 51.22\% \\
    \cline{1-4}
    RTN & 36.59\% & 33.54\% & 46.34\% \\
    AWQ & 35.98\% & 31.71\% & 50.61\% \\
    SmoothQuant+ & \pmb{35.98\%} & \pmb{37.80\%} & \pmb{53.05\%} \\
    \cline{1-4}
  \end{tabular*}
  }
\end{table}

\subsubsection{Multilingual Evaluation}
For further investigation, we use the BabelCode framework to conduct multilingual evaluation. We report results for Python, JAVA, GO, C++ in Table \ref{tbl:humaneval_multilingual}. It can be seen that SmoothQuant+ outperforms the FP16 model by 0.6\% in the multilingual setting on average, especially 3.05\% in Python and 5.59\% in JAVA.

\begin{table*}[!ht]
\small
  \caption{Code Llama-34B multilingual HumanEval pass@1. The BabelCode\cite{orlanski2023measuring} framework which converts the HumanEval (Python) set into other programming languages, including Python, JAVA, Go, C++, is used for multilingual evaluation.}
  \label{tbl:humaneval_multilingual}
  \setlength{\tabcolsep}{8mm}{
  \begin{tabular*}{\textwidth}{lccccc}
    \cline{1-6}
    HumanEval$\uparrow$ & Python & JAVA & GO & C++ & Average \\
    \cline{1-6}
    FP16 & 51.22\% & 38.51\% & \pmb{26.71\%} & \pmb{45.34\%} & 40.45\% \\
    \cline{1-6}
    SmoothQuant+ & \pmb{54.27\%} & \pmb{44.10\%} & 24.22\% & 41.62\% & \pmb{41.05\%} \\
    \cline{1-6}
  \end{tabular*}
  }
\end{table*}
\subsection{Throughput and Latency Evaluation}
\begin{figure*}[!ht]
 \centering
 \includegraphics[height=6cm,keepaspectratio]{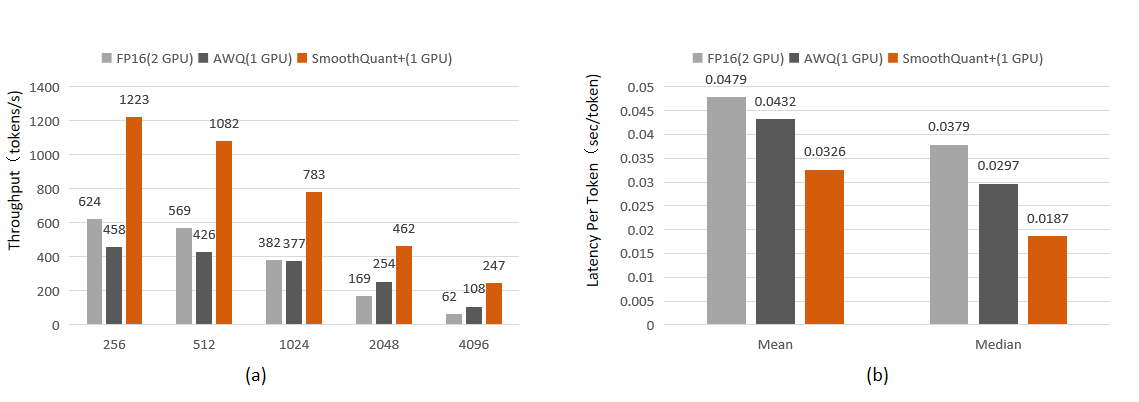}
 \caption {Code Llama-34B, comparison of inference throughput and latency on vLLM engine.}
 \label{fgr:throughput_latency}
\end{figure*}

We use Poisson process to synthesize the request arrival times. By setting different input and output lengths, the ultimate throughput of the model under various given context lengths is shown in Figure \ref{fgr:throughput_latency}(a). Furthermore, the real access traffic in the online environment is used as the test data for simulation testing. The content requested by the user and the interval between requests are consistent with those online. The delay of a single token is calculated in Figure \ref{fgr:throughput_latency}(b). It can be seen that the performance of an AWQ deployed by one NVIDIA A100 40GB PCIe is weaker than models in FP16 deployed on two NVIDIA A100 40GB PCIe GPUs in terms of throughput and latency. By SmoothQuant+, the Code Llama-34B model can be quantified and deployed on a single A100 40GB GPU, with lossless accuracy and a throughput increase of 1.9-4.0 times that of the model in FP16  deployed on two A100 40GB GPUs. The latency per token is only 68\% of the model in FP16. This is the state-of-the-art 4-bit weight-only quantization as we know.

\subsection{Ablation}
\subsubsection{Sensitivity Analysis of Calibration Sets}

Our method uses the calibration set as the input of the model to calculate the maximum activation values by channel when smoothing the model, and also uses the calibration set as input to calculate quantization losses to determine the smooth strength $\alpha$. The reason why the problem descriptions in HumanEval\cite{chen2021evaluating} are chosen as the calibration set is because we find that using SmoothQuant+, the final HumanEval accuracy is sensitive to the calibration set. This can be seen from Table \ref{tbl:diff_calib_set} that a subset of the Pile\cite{gao2020pile} validation set, a subset of the C4 dataset\cite{raffel2020exploring}, and the problem descriptions in HumanEval are used as calibration sets during testing. As a calibration set, the problem descriptions in HumanEval achieve the highest HumanEval pass@1 among Code Llama-7B, 13B, and 34B. The results of the other two sets are not ideal.

\begin{table}
\small
  \caption{For Code Llama family, the impact of different calibration sets on SmoothQuant+, searching step=0.05. SmoothQuant+ performs 4-bit weight quantization and uses the problem descriptions in HumanEval as the calibration set to obtain the best HumanEval pass@1.}
  \label{tbl:diff_calib_set}
  \setlength{\tabcolsep}{4mm}{
  \begin{tabular*}{\textwidth}{lccc}
    \cline{1-4}
    HumanEval$\uparrow$ & 7B & 13B & 34B \\
    \cline{1-4}
    Pile & 28.05\% & 32.32\% & 50\% \\
    C4 & 31.71\% & 32.32\% & 45.12\% \\
    HumanEval & \pmb{35.98\%} & \pmb{37.80\%} & \pmb{53.05\%} \\
    \cline{1-4}
  \end{tabular*}
  }
\end{table}

\subsubsection{Step Selection in Searching}
When smoothing, it is crucial to choose the appropriate smoothing strength $\alpha$. We search between 0 and 1 at certain step intervals to obtain the smoothing strength that minimizes the quantization loss of the entire model. We try different search intervals and find that step=0.05 can obtain the best results of 4-bit weight quantization for Code Llama family. The quantized model is lossless in accuracy. When step=0.01, the results is not as good as step=0.05. However, it still has postive effect of using the searched smoothing strength to smooth the model, its performance is basically better than RTN, and on the same level as AWQ. This shows that SmoothQuant+ is insensitive to step, as shown in Table \ref{tbl:diff_searching_step}. Finally, we recommend using step=0.05. The reason why different steps sometimes search for different smoothing strength is that when the quantization loss is low, the HumanEval accuracy after model quantization is usually high. But there is inevitably slight fluctuations in accuracy because the difference in loss is too small, usually the fourth or fifth digits after the decimal point.

\begin{table*}
\small
  \caption{Using the problem descriptions in HumanEval as the calibration set, SmoothQuant+ is not sensitive to the searching step. The loss is the quantization loss of the Code Llama family.}
  \label{tbl:diff_searching_step}
  \setlength{\tabcolsep}{6mm}{
  \begin{tabular*}{\textwidth}{lccc}
    \cline{1-4}
    HumanEval$\uparrow$ / (loss) & 7B & 13B & 34B \\
    \cline{1-4}
    FP16 & 35.98\% & 35.98\% & 51.22\% \\
    RTN & 36.59\% & 33.54\% & 46.34\% \\
    AWQ & 35.98\% & 31.71\% & 50.61\% \\
    SmoothQuant+(step=0.05) & \pmb{35.98}\%/(0.00449) & \pmb{37.80}\%/(0.00399) & \pmb{53.05}\%/(0.00363) \\
    SmoothQuant+(step=0.01) & \pmb{35.98\%}/(0.00449) & 33.54\%/(0.00393) & 48.17\%/(0.00354) \\
    \cline{1-4}
  \end{tabular*}
  }
\end{table*}

\section{Related Work}
LLMs have achieved astonishing results in various tasks\cite{radford2019language,brown2020language,wei2021finetuned,openai2023gpt}. But the accompanying problems are the increasing number of model parameters and the increasing demand for memory footprint and computing resources. How to efficiently deploy LLMs has become a concern in both academia and industry. In order to reduce computational and storage costs, accelerate inference speed, model compression is generally used. There are many model compression methods that have achieved good results in convolutional networks such as sparsity, quantization, distillation, and low rank decomposition. Quantization is the most direct method for compressing models and accelerating inference, but the large number of parameters and different transformers based structures of LLMs determine their specificity in optimizing them.

Classic quantization methods include MinMax quantization\cite{jacob2018quantization,wu2020integer}, cross-entropy quantization\cite{wu2020integer}, percentile quantization\cite{wu2020integer}, LSQ\cite{esser2019learned}, PACT\cite{choi2018pact}, etc. They are divided into QAT and PTQ according to whether training is required when quantizing. Because QAT requires model training during the quantization, and training LLMs requires a large amount of computing resources and time, researchers have payed more attention to the PTQ of LLMs.

Most of the original quantization methods are aimed at convolutional neural networks. After the number of parameters exceeds 6.7B, systematic outliers appear in the activation\cite{dettmers2022llm} which cause large quantization loss when using original quantization methods. As a result, there have been many studies\cite{frantar2022gptq,lin2023awq,lee2023owq,yao2022zeroquant,dettmers2022llm,xiao2023smoothquant,dettmers2023spqr} on PTQ for LLMs. 8-bit PTQ has currently achieved good results, but with 4-bit PTQ weight quantization, the accuracy of the model generally decreases, and the inference acceleration is not obvious. SmoothQuant+ solves the above problems.

The closest works to SmoothQuant+ are SmoothQuant\cite{xiao2023smoothquant} and AWQ\cite{lin2023awq}. SmoothQuant+ is better than AWQ in both accuracy and inference speed as shown in Table \ref{tbl:comparison_differert_quant}. 

\begin{table*}
\small
  \caption{Comparison of SmoothQuant+ , SmoothQuant and AWQ. Accuracy $\checkmark$ indicates lossless accuracy. Efficiency is tested on vLLM, and $\checkmark$ indicates that the throughput and latency deployed on one A100 40GB GPU are better than the model deployed on two GPUs in FP16.}
  \label{tbl:comparison_differert_quant}
  \setlength{\tabcolsep}{7mm}{
  \begin{tabular*}{\textwidth}{ccccc}
    \cline{1-5}
    Quantization method & Weight bits & Activation bits & Accuracy & Efficiency \\
    \cline{1-5}
    SmoothQuant & 8 & 8 & \checkmark & - \\
    AWQ & 4 & 16 & $\times$ & $\times$  \\
    SmoothQuant+ & 4 & 16 & \checkmark & \checkmark  \\
    \cline{1-5}
  \end{tabular*}
  }
\end{table*}

SmoothQuant+ uses the same smoothing method as SmoothQuant to smooth the activation and weights of the model except that it starts with a different starting point. SmoothQuant+ first smooths the model and then quantizes the weights in 4-bit. During this process, the activation is not quantized. However, smoothing activation is still required, mainly because the activation outliers may cause significant quantization losses, and quantization losses can be reduced through smoothing activation. SmoothQuant needs to perform 8-bit quantization on both activations and weights after smoothing the model, whose weights are only per-tensor quantized. We consider that SmoothQuant+ which is a 4-bit group-wise quantification is a further extension of SmoothQuant. AWQ also performs 4-bit weight quantization like SmoothQuant+, but the different purpose of smoothing results in different implementation. The AWQ’s purpose of smoothing activation is to protect the salient weights. When selecting the importance factor by channel in activation, the mean is selected instead of the maximum. AWQ searches for the hyper-parameter $\alpha$ layer by layer, and the objective is to minimize the quantization loss of the specified layer. When AWQ calculates this quantization loss, it doesn't consider the impact of the quantization of the previous layers on the subsequent layers which causes an error accumulation. We believe that it is inappropriate to search for hyper-parameter layer by layer without considering error accumulation, which results in a reduction in accuracy. Moreover, as the model size increases and the number of linear layers increases, the searching time will increase significantly.

\section{Conclusion}
We propose SmoothQuant+, a simple and effective post-training quantization method to enable lossless 4-bit group-wise weight quantization. SmoothQuant+, a tribute and extension to SmoothQuant, is proposed based on the fact that large quantization losses are caused by activation outliers in LLMs and smoothing activation can reduce quantization loss. We implemented this algorithm on vLLM, enabling the Code Llama-34B model to be quantified and deployed on a NVIDIA A100 40GB GPU with lossless accuracy. The throughput increase is 1.9-4.0 times that of the model in FP16 deployed on two NVIDIA A100 40GB GPUs, and the latency per token is only 68\% of the model in FP16. This is the best result we know of 4-bit weight quantization.



\balance


\bibliography{rsc} 
\bibliographystyle{rsc} 

\end{document}